\documentclass[lettersize,journal]{IEEEtran}
\usepackage{amsmath,amsfonts}
\usepackage{algorithmic}
\usepackage{algorithm}
\usepackage{array}
\usepackage[caption=false,font=normalsize,labelfont=sf,textfont=sf]{subfig}
\usepackage{textcomp}
\usepackage{stfloats}
\usepackage{url}
\usepackage{verbatim}
\usepackage{graphicx}
\usepackage{cite}
\usepackage[colorlinks,
linkcolor=blue,
anchorcolor=blue,
citecolor=blue,backref]{hyperref}
\usepackage[all]{hypcap}
\usepackage{bbding}
\usepackage{multirow}
\usepackage{ulem}
\usepackage{float}
\usepackage{soul, color, xcolor}

\begin{document}
	
	\title{Interpretable Deep Learning Paradigm for Airborne Transient Electromagnetic Inversion}
	
	\author{Shuang Wang, Xuben Wang, Fei Deng, Peifan Jiang and Lifeng Mao\vspace{-5mm}    
		\thanks{This work was supported by the Deep Earth National Science and Technology Major Project under Grant 2024ZD1002905 and National Key Research and Development Program of China under Grant 2023YFB3905004. (Corresponding author: Xuben Wang.)}
		\thanks{S. Wang, X. Wang, P. Jiang and L. Mao are with the Key Laboratory of Earth Exploration and Information Techniques, Ministry of Education, College of Geophysics, Chengdu University of Technology, Chengdu 610059, China (e-mail: wangs@stu.cdut.edu.cn; wxb@cdut.edu.cn; jpeifan@qq.com; maolifeng07@cdut.cn;).
			
		Fei Deng is with the College of Computer Science and Cyber Security, Chengdu University of Technology, Chengdu 610059, China (e-mail: dengfei@cdut.edu.cn).}}
	
	\markboth{Journal of \LaTeX\ Class Files,~Vol.~14, No.~8, August~2021}%
	{Shell \MakeLowercase{\textit{et al.}}: A Sample Article Using IEEEtran.cls for IEEE Journals}
	
	
	\maketitle
	
	\begin{abstract}
The extraction of geoelectric structural information from airborne transient electromagnetic (ATEM) data primarily involves data processing and inversion. Conventional methods rely on empirical parameter selection, making it difficult to process complex field data with high noise levels. Additionally, inversion computations are time-consuming and often suffer from multiple local minima. Existing deep learning-based approaches separate the data processing steps, where independently trained denoising networks struggle to ensure the reliability of subsequent inversions. Moreover, end-to-end networks lack interpretability. To address these issues, a unified and interpretable deep learning inversion paradigm based on disentangled representation learning is proposed. The network explicitly decomposes noisy data into noise and signal factors, completing the entire data processing workflow based on the signal factors, which makes the network more reliable and interpretable. Furthermore, physical constraints are incorporated into the learning process to enhance the physical consistency and reliability of the inversion results. The inversion results on field data demonstrate that the method can directly use noisy data to accurately reconstruct the subsurface electrical structure, thereby establishing a unified, interpretable, and physically constrained inversion paradigm for ATEM data processing.
	\end{abstract}
	
	
	\section{Introduction}
	\IEEEPARstart{t}{he} airborne transient electromagnetic (ATEM) method is a widely used technique in electromagnetic detection and a current focal point of research in geological exploration. It involves a transmitting loop that generates a time-varying current during flight, while receiving coils capture the induced secondary electromagnetic field produced by subsurface media to map underground structures \cite{palacky1991airborne}. By mounting both the transmitting and receiving systems on the aircraft, ATEM facilitates rapid coverage of extensive and difficult-to-reach survey areas. It has been widely applied in geohazards investigations \cite{damhuis2020identification,malehmir2016near}, groundwater resource exploration \cite{ball2020high,minsley2021airborne}, mineral prospecting \cite{kone2021geophysical,okada2021historical}, and geological mapping \cite{dzikunoo2020new,wong2020interpretation}.
	
	The primary process of extracting geoelectric structural information from ATEM data involves data processing and inversion \cite{wu2022deep}. The core task of data processing, beyond steps such as system response effect removal, is denoising. Due to the dynamic nature of data acquisition, complex noise conditions, long observation distances, and relatively weak signal strength, ATEM data processing presents significant challenges \cite{wu2019removal}. Traditional physics-based methods, including principal component analysis \cite{asten2009use}, stationary wavelet transform \cite{li2017electromagnetic}, singular value decomposition \cite{reninger2011singular}, and ensemble empirical mode decomposition \cite{liu2017application}, typically rely on the selection of empirical parameters. This limitation not only demands significant operator expertise but also restricts their effectiveness primarily to low-noise scenarios. Current field applications reveal notable performance degradation when these conventional approaches encounter complex geological environments with elevated noise contamination levels, particularly in challenging survey conditions involving strong electromagnetic interference and low signal-to-noise ratios \cite{wu2020removal}.
	
	Inversion directly reflects the geoelectric structure. Conventional inversion methods, such as the Occam inversion method \cite{constable1987occam,vallee2009application} and some regularized approaches \cite{siemon2009laterally,vallee2009inversion,vignoli2015sharp,christensen2017voxel}, rely on forward modeling to iteratively refine the inversion results \cite{bai20211d}. This process is computationally expensive, with costs increasing geometrically as the number of observations grows, making it prohibitively expensive for large-scale datasets. Moreover, due to noise and the inherent non-uniqueness of inverse problems, regularization methods require relatively accurate prior information (initial models) of the subsurface structure, often encountering challenges with multiple local minima \cite{cheng2024transient}. Probabilistic inversion methods, such as Bayesian-based approaches, provide a probability distribution of resistivity values, transforming the inversion problem into one of determining the probability distribution of inversion parameters \cite{cheng2024transient}. While this method is suitable for highly nonlinear problems, however, its computational cost is high and increases exponentially with the number of model parameters \cite{wu2023fast}.
	\begin{figure*}[!t]
		\centering
		\includegraphics[width=6in]{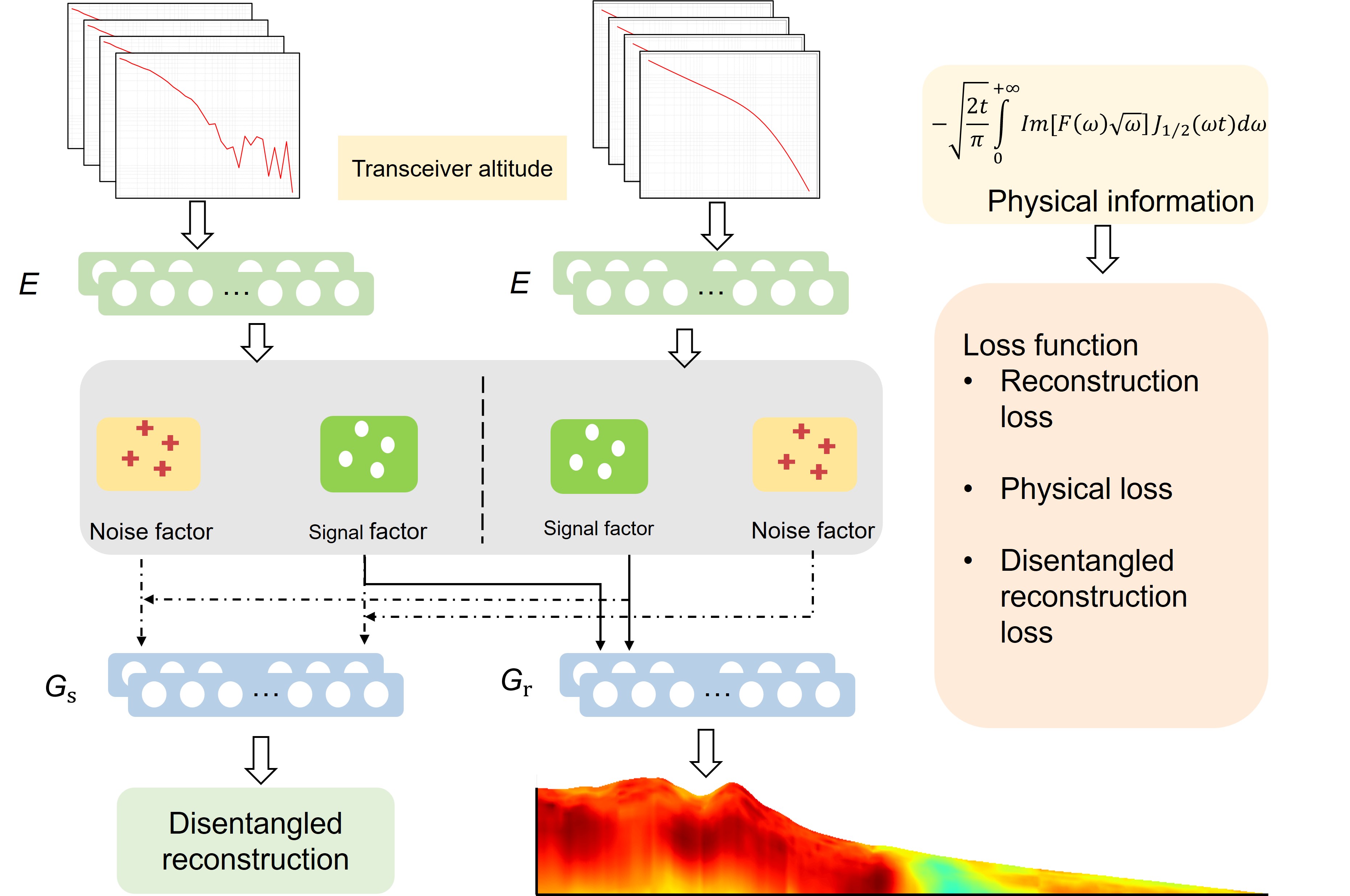}
		\caption{The overall framework, the data is encoded by the encoder $E$ into signal and noise factors. The inversion decoder $G_r$ decodes the signal factors to obtain accurate inversion results. $G_s$ is used to decode the combination of signal and noise factors into the signal, ensuring the correct disentangled representation. The loss function incorporates physical information to impose physical constraints.}
		\label{1}
	\end{figure*}
	
	With the advancement of deep learning, its application in denoising and inversion of geophysical data has attracted significant attention as a means of overcoming the limitations of traditional methods \cite{wang2026dremnet,wang2026rapid,wang2026aotem,wang2025openem,li2026dmtnet}. For example, denoising autoencoders have been used to remove multi-source noise from data \cite{wu2020removal}. One-dimensional signals have been transformed into two-dimensional sequences for transient electromagnetic denoising using image denoising networks \cite{chen2020temdnet}. Wu \textit{et al.} \cite{wu2021noising} employed a Long Short-Term Memory (LSTM) network combined with an autoencoder structure to denoise data. Liu \textit{et al.} \cite{liu2024multi} proposed a Transformer-based multi-task learning network architecture for denoising data. Cheng \textit{et al.} \cite{cheng2025pc} utilized parallel convolution to extract multi-scale features from input data and then applied a bidirectional Long Short-Term Memory (BiLSTM) network to establish complex mapping relationships between features and clean data for transient electromagnetic denoising. Additionally, data-driven convolutional neural network (CNN)-based inversion has been employed \cite{wu2021convolutional}, LSTM utilized for fast inversion \cite{wu2022instantaneous}.  Wu \textit{et al.} \cite{wu2024physics} developed a hybrid deep learning-based method that integrates deep learning-based forward modeling losses into the loss function to yield more consistent inversion results. Kang \textit{et al.} \cite{kang2024deep} designed a deep learning-based inversion process, while Zhang \textit{et al.} \cite{zhang2024airborne} introduced a multi-input, single-output double-hidden wavelet neural network for ATEM data imaging. Cheng \textit{et al.} \cite{cheng2024multi} proposed a TEM inversion algorithm based on a multi-scale expanded convolutional neural network, aiming to fully exploit the multi-scale information in the data to improve inversion reliability. Furthermore, Cheng \textit{et al.} \cite{cheng2024transient} proposed an inversion method using CNN-LightGBM. 
	
	However, these deep learning approaches treat denoising and inversion as separate processes, thereby creating a distinct boundary between them. This separation necessitates the training of two independent networks—one for denoising and another for inversion. Since inversion is a downstream task that heavily depends on the accuracy of the denoising results, it typically evaluates the reliability of the estimated model by calculating the deviation between the model's response and the processed data, but it cannot simultaneously assess the "purity" of the processed data. As a result, if residual interference remains in the processed data, the reliability of the inversion is compromised \cite{wu2022deep}. Moreover, training two separate networks requires significant computational resources and time. Although multi-task learning attempts to jointly optimize denoising and inversion within a single framework, its shared latent representations typically entangle inversion-related signal features with noise-related information. This mixed representation inevitably introduces interference during the decoding process, limiting inversion accuracy under complex noise conditions. More importantly, existing network architectures are generally black-box mappings that lack interpretation of intermediate representations and fail to explicitly model the separation process between noise and signal. As a result, current end-to-end networks produce outputs that are inherently non-interpretable.
	
	To address the aforementioned challenges, we propose a unified and interpretable deep learning inversion paradigm based on disentangled representation learning. The network explicitly decouples noisy data into noise and signal factors, using the signal factors for inversion, thereby unifying the denoising and inversion of field data. This approach organically integrates denoising and inversion into a single workflow, eliminating error propagation and the need for manual preprocessing.Moreover, by explicitly modeling the separation process of noise and signal, the network attains improved interpretability.  The network also incorporates physical information as constraints and completes the entire data processing pipeline based on the signal factors, rendering the network's results more reliable. The inversion results on field data demonstrate that our model can accurately and rapidly yield the subsurface electrical structure in a unified manner, without requiring manual data preprocessing.The contributions of this paper are as follows:
	
	1. A unified and interpretable deep learning inversion paradigm is constructed based on disentangled representation learning, which resolves the issues associated with the stepwise processing of denoising and inversion, thereby enabling accurate inversion.
	
	2. By explicitly modeling and disentangling signal and noise factors, the interpretability of the model is enhanced.
	
	3. The network incorporates physical information as constraints and completes the entire data processing pipeline based on the signal factors, thereby making the network's processing results more reliable.
	\begin{figure*}[!t]
		\centering
		\includegraphics[width=6in]{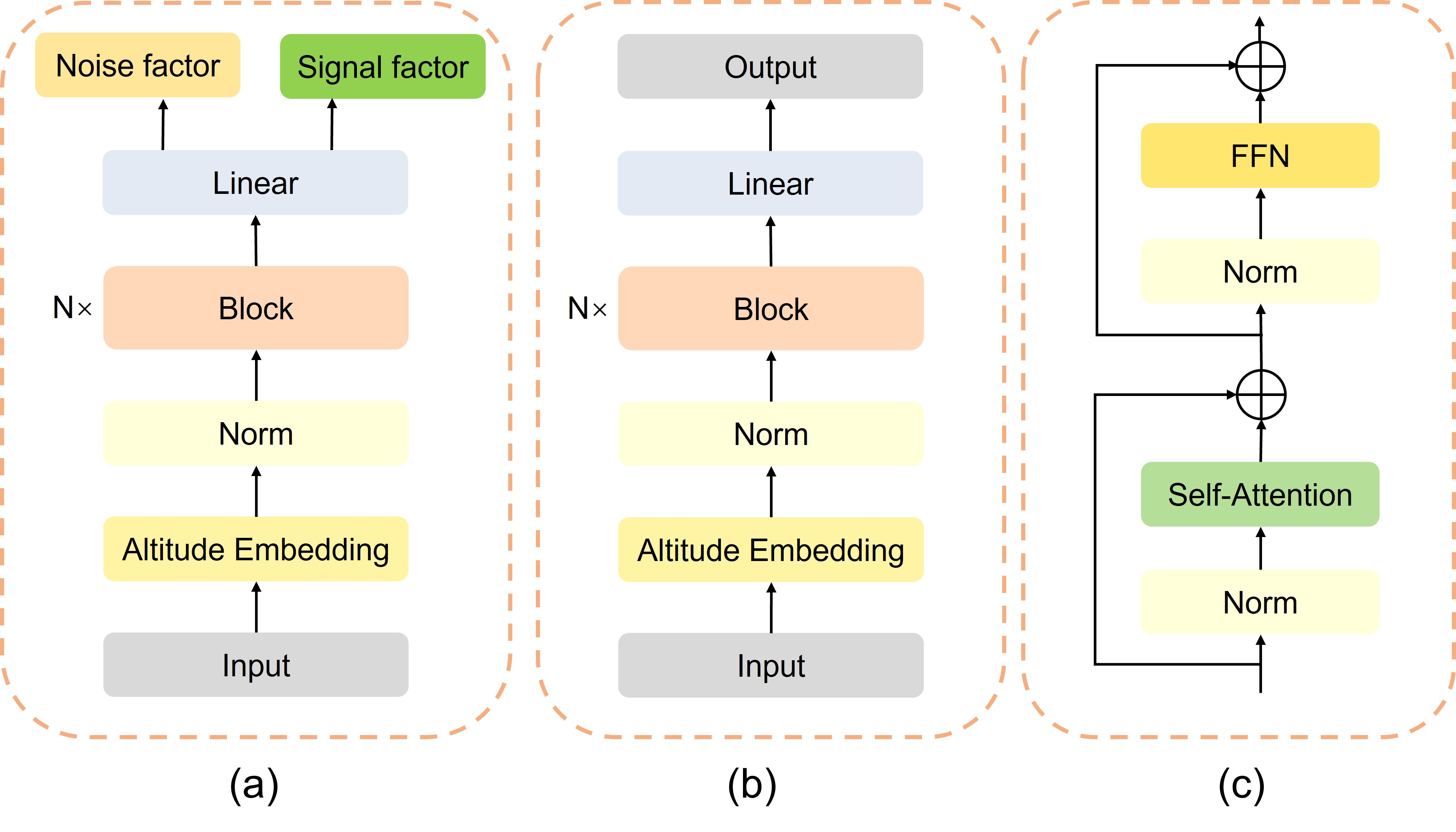}
		\caption{(a) represents the architecture of the encoder E, (b) represents the architectures of the inversion decoder Gr and the data decoder Gs, and (c) shows the stacked block structure within the encoder and decoder.}
		\label{2}
	\end{figure*}
	
	\section{Method}
	Existing deep learning methods are typically designed for specific stages of the data processing pipeline, such as denoising or inversion, without integrating both processes into a unified framework. However, because inversion inherently depends on denoised data, integrating these two stages can streamline the entire workflow.
	\begin{figure*}[!t]
		\centering
		\includegraphics[width=6in]{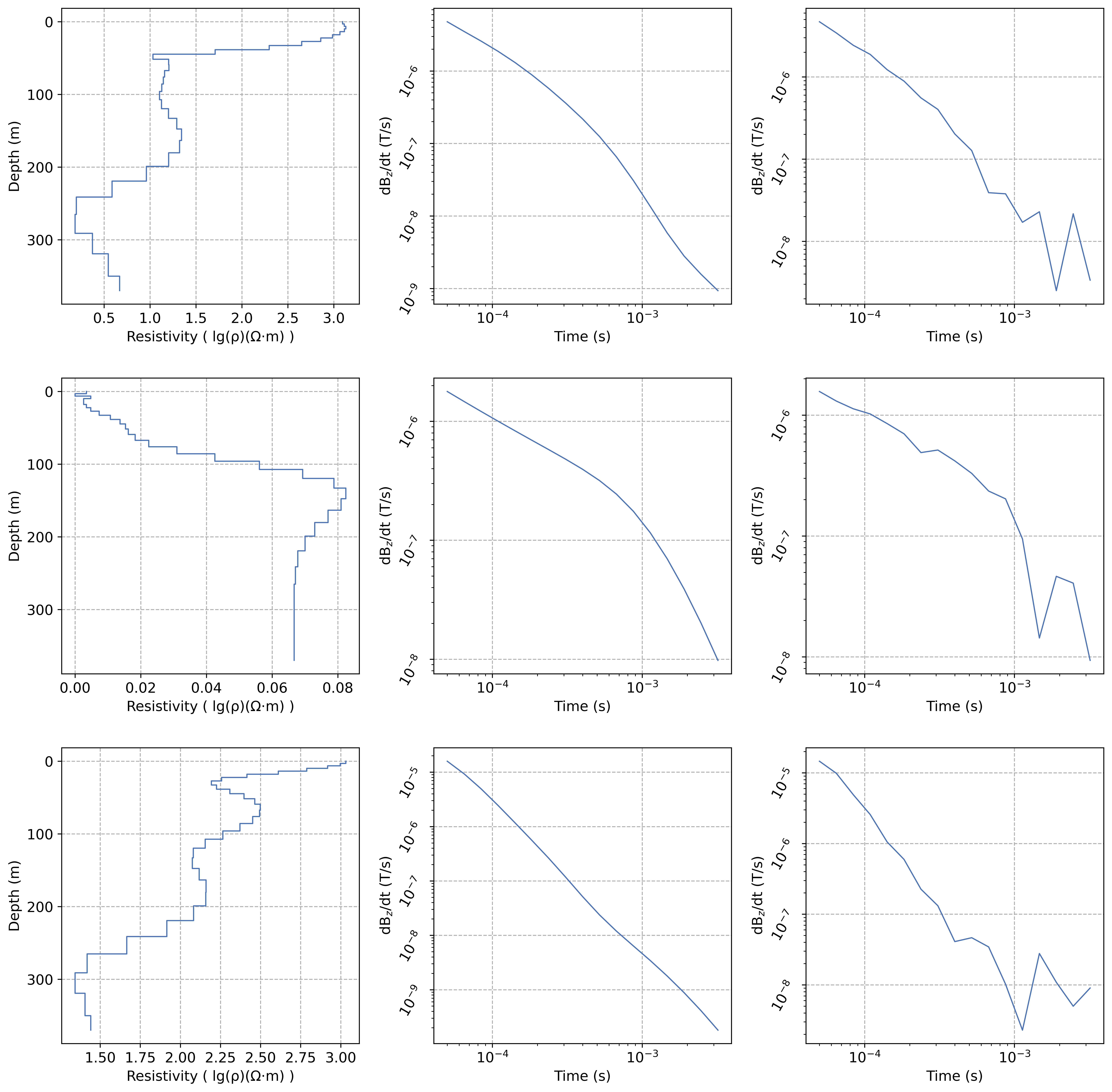}
		\caption{The 30-layer RMD model (on the left), forward response (in the middle), and the data after noise addition (on the right).}
		\label{3}
	\end{figure*}
	
	Disentangled representation learning seeks to model key representations within data by separating task-relevant factors from task-irrelevant ones \cite{wang2024disentangled}. In the context of denoising and inversion, the useful data representation within the signal corresponds to task-relevant factors, while noise represents task-irrelevant factors. Since the task-relevant factors for both denoising and inversion are the same—referred to as the signal factors—we leverage disentangled representation learning to decouple data into signal and noise factors, performing inversion based on the signal factors. This approach seamlessly integrates denoising and inversion into a unified framework. Additionally, we employ a data decoder to ensure that the encoder correctly disentangles the data. The overall framework is depicted in Fig. \ref{1}.
	
	The network comprises an encoder \( E \), an inversion decoder \( G_r \), and a data decoder \( G_s \). The encoder \( E \) encodes the input data into signal and noise factors, while the inversion decoder \( G_r \) decodes the signal factors to generate accurate inversion results. The data decoder \( G_s \) decodes the combined signal and noise factors into the signal, ensuring the correctness of the disentangled representation.
	
	Given the collected data \( x \), the encoder \( E \) encodes it into a pair of disentangled representations:
		\begin{equation}
			Z_s, Z_n = E(x)
		\end{equation}
	Here, \( Z_s \) represents the signal factors, and \( Z_n \) represents the noise factors. A mutual information upper bound estimator, CLUB \cite{cheng2020club}, is employed to ensure that the distributions of \( Z_s \) and \( Z_n \) are fully separated, as the distributions of the signal and noise are entirely distinct.
	\begin{figure*}[!t]
		\centering
		\includegraphics[width=6in]{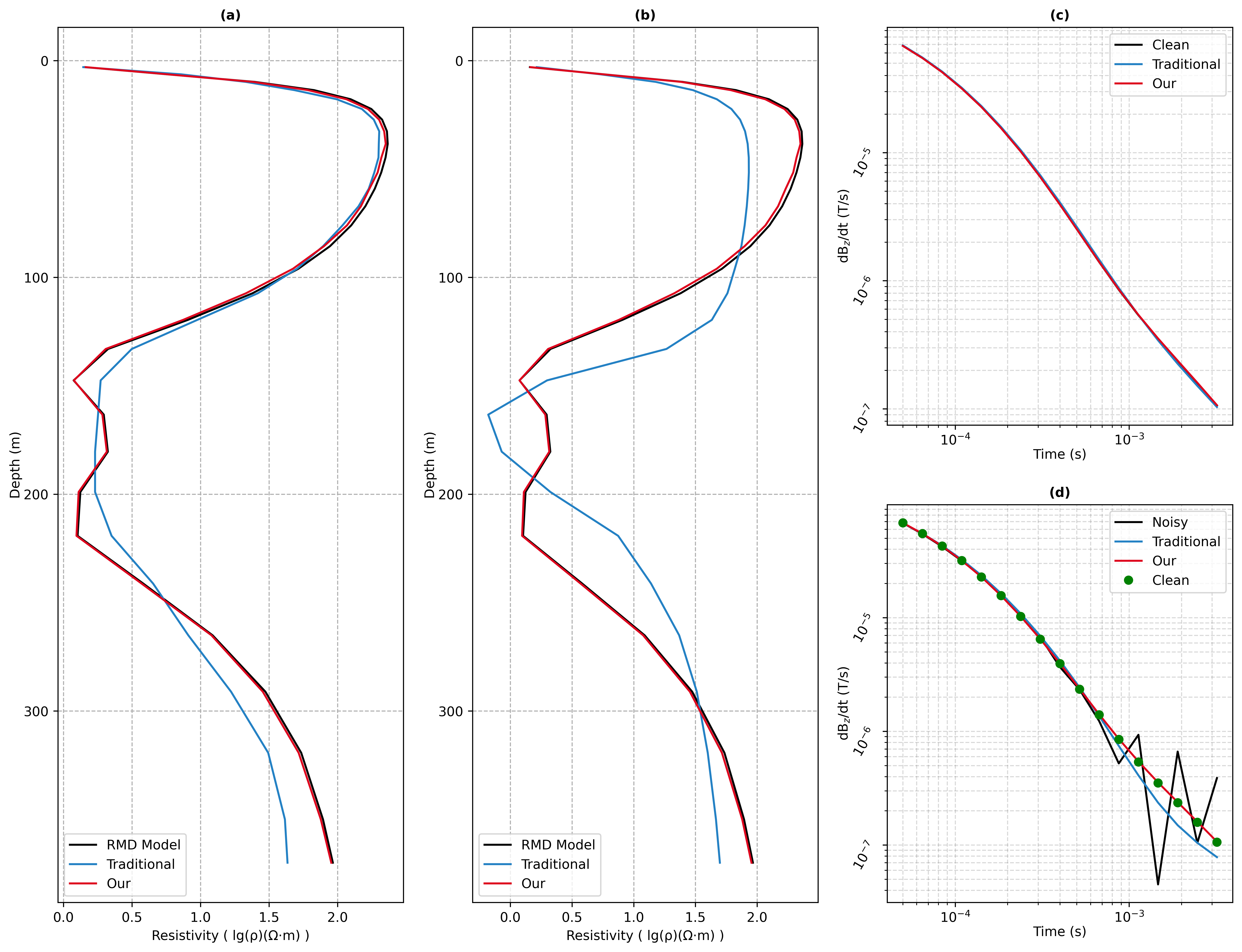}
		\caption{(a) Inversion results using clean signals, (b) Inversion results using noisy signals, (c) Forward modeling response results using the inversion model from (a), (d) Forward modeling response results using the inversion model from (b)}
		\label{4}
	\end{figure*}
	\begin{figure*}[!t]
		\centering
		\includegraphics[width=6in]{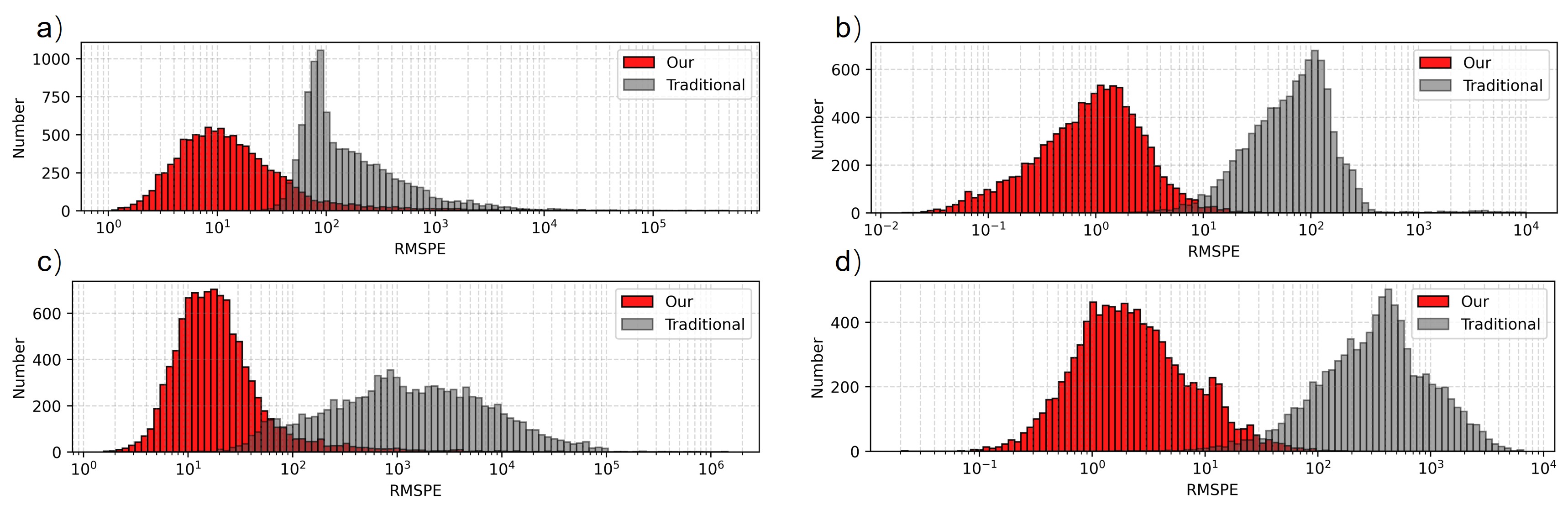}
		\caption{(a) RMSPE statistics of the inversion results for clean signals across the entire test set, (b) RMSPE statistics computed from the forward modeling response of the inversion results in (a) and the clean signals, (c) RMSPE statistics of the inversion results for noisy signals across the entire test set, (d) RMSPE statistics computed from the forward modeling response of the inversion results in (c) and the clean signals.}
		\label{5}
	\end{figure*}
	
	CLUB is a novel contrastive log-ratio upper bound (CLUB) of mutual information with effectiveness and generality in the real-world mutual information minimization experiments. We adopt the CLUB mutual information estimator to quantify the discrepancy between two latent vectors. To disentangle the signal and noise factors within the input data, we minimize the mutual information between $Z_s$ and $Z_n$ using CLUB, as shown in equation 2.
	\begin{equation}
		\begin{aligned}
		L_{\text{club}}(Z_s, Z_n) &= \mathbb{E}_{p(Z_s,Z_n)}[\log p(Z_n \mid Z_s)] \\ &- \mathbb{E}_{p(Z_s)} \mathbb{E}_{p(Z_n)}[\log p(Z_n \mid Z_s)]
	    \end{aligned}
	\end{equation}
	The conditional probability distribution $p(Z_n \mid Z_s)$ about $Z_n$ and $Z_s$, which is unknown and not computable. To obtain this distribution, a parameterized variational network $q_\theta(Z_n \mid Z_s)$ is used to approximate $p(Z_n \mid Z_s)$. The mutual information upper bound estimator is trained by maximizing the log-likelihood of $q_\theta(Z_n \mid Z_s)$. By maximizing this log-likelihood, an approximate distribution of $p(Z_n \mid Z_s)$ is learned through the network $q_\theta(Z_n \mid Z_s)$. Subsequently, $q_\theta(Z_n \mid Z_s)$ is used in place of $p(Z_n \mid Z_s)$ in equation 2 to obtain the upper bound of mutual information between $Z_s$ and $Z_n$.
	
	During training, the encoder \( E \) is used to encode both noisy signals \( n \) and clean signals \( s \). The network is expected to correctly obtain the disentangled representations \( Z_s^1, Z_n^1 = E(n) \) and \( Z_s^2, Z_n^2 = E(s) \). The signal factors representations, \( Z_s^1 \) and \( Z_s^2 \), should follow the same distribution, indicating they represent similar signal factors. In contrast, the noise factors differ, as the clean signal contains no noise. The noise factors are then swapped between the two representations, and the data decoder is employed to decode them. Decoding \( (Z_s^1, Z_n^2) \) yields the clean signal, while decoding \( (Z_s^2, Z_n^1) \) yields the noisy signal, demonstrating that the data has been fully disentangled. Finally, the inversion decoder \( G_r \) is used to decode the signal factors and obtain the inversion result.
		\begin{equation}
			m = G_r(Z_s^1) = G_r(Z_s^2)
		\end{equation}
	
	To address the challenge of data-driven networks relying on the completeness of training data, which necessitates that the training set adequately represent the variations in geophysical properties \cite{wu2024physics}, we incorporate physical information to guide and drive the network training process. This approach ensures that the model avoids biased interpretations of subsurface structures when applied to field observations. Given the resistivity model \( m \), its forward response is:
		\begin{equation}
			s = O(m)
		\end{equation}
	Here, \( O \) represents the forward operator, and its computed response is as follows: 
	
	Using the step waveform as an example, the current function is given by:
		\begin{equation} 
			\begin{cases}
				I(t) = I_0, t \le 0\\
				I(t) = 0, t > 0
			\end{cases}
		\end{equation}
	The response result is given by:
		\begin{equation}  
			f(t) = \frac{1}{2\pi} \int_{-\infty}^{+\infty} F(\omega) e^{i\omega t} d\omega 
		\end{equation}
	The term \( e^{i\omega t} \) is the time-harmonic factors, where \( \omega \) represents the angular frequency, and \( F(\omega) \) is the frequency domain response. This can be further derived as \cite{yin2013full}:  
		\begin{equation} 
			f(t) = -\sqrt{\frac{2t}{\pi}} \int_0^{+\infty} \text{Im}\left[ F(\omega) \sqrt{\omega} \right] J_{\frac{1}{2}}(\omega t) d\omega
		\end{equation}
	Where $J_{\frac{1}{2}}(\omega t)$ is the half-integer order Bessel function. The response of an arbitrary emitted waveform is given by:  
		\begin{equation} 
			\frac{dB}{dt} = \frac{dI(t)}{dt} * f(t)
		\end{equation}
	Where, \( t \) represents the given time, and \( I(t) \) is the emitted current. The forward operator is applied to the inversion result $m$ to obtain the physical constraint loss:
		\begin{equation}
			L_{physic}=||O(m_{Z_s^1}) - s||^2 + ||O(m_{Z_s^2}) - s||^2
		\end{equation}
	Where, $m_{Z_s^1}$ denotes the inversion result obtained using $Z_s^1$, while $m_{Z_s^2}$ represents the inversion result derived from $Z_s^2$.
	The overall loss function is:
		\begin{equation}
			L=L_{physic}+L_{clean}+L_{noise}+L_{rho}+L_{club}
		\end{equation}
		\begin{equation}
			L_{clean} = ||G_s(Z_s^2, Z_n^2) - s||^2 + ||G_s(Z_s^1, Z_n^2) - s||^2
		\end{equation}
		\begin{equation}
			L_{noise} = ||G_s(Z_s^2, Z_n^1) - n||^2 + ||G_s(Z_s^1, Z_n^1) - n||^2
		\end{equation}
		\begin{equation}
			L_{rho} = ||m_{Z_s^1}-m_l||^2 + ||m_{Z_s^2}-m_l||^2
		\end{equation}
	Where \( m_l \) is the true resistivity model.
	\begin{figure*}[!t]
		\centering
		\includegraphics[width=6in]{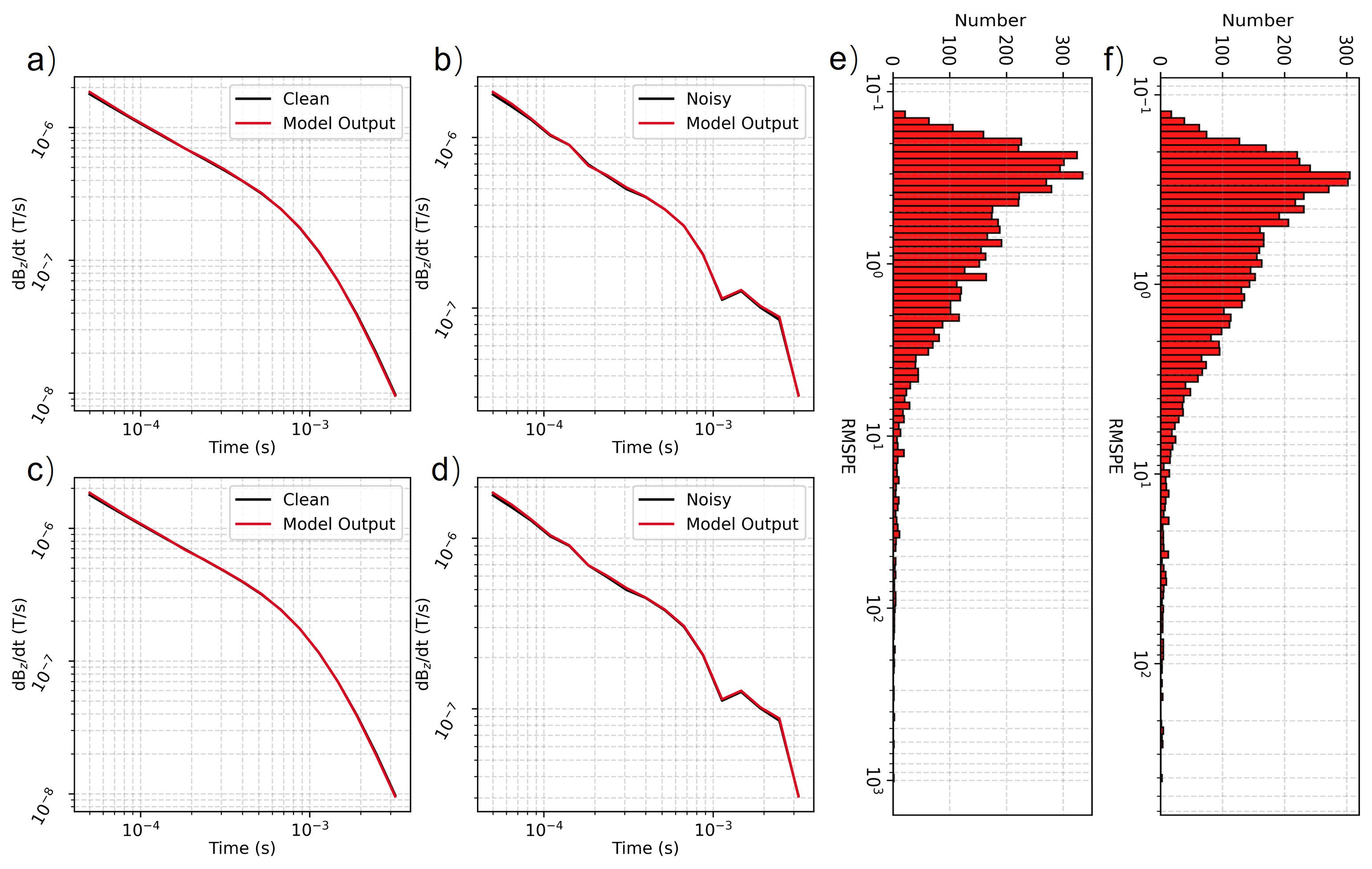}
		\caption{The clean signal \( s \) is decoupled into the signal factors \( Z_s^1 \) and the noise factors \( Z_n^1 \), while the noisy signal \( n \) is decoupled into the signal factors \( Z_s^2 \) and the noise factors \( Z_n^2 \). (a) The clean signal decoded using \( (Z_s^1, Z_n^1) \), (b) the noisy signal decoded using \( (Z_s^2, Z_n^2) \), (c) the clean signal decoded using \( (Z_s^2, Z_n^1) \), (d) the noisy signal decoded using \( (Z_s^1, Z_n^2) \). (e) The RMSPE statistics of the clean signal decoded from \( (Z_s^1, Z_n^1) \) on the test set, (f) The RMSPE statistics of the clean signal decoded from \( (Z_s^2, Z_n^1) \) on the test set.}
		\label{6}
	\end{figure*}
	
	Since variations in transceiver height also affect the data, we further embedded the transceiver height into the network. Our focus is on the entire data processing pipeline rather than the network architecture itself. Therefore, the network structure follows the GPT-2 style \cite{lagler2013gpt2}, with the encoder depicted in Fig. \ref{2}(a). The input is embedded, while the height information is simultaneously linearly embedded, followed by normalization via a normalization layer. Subsequently, the data is processed by $N$ Blocks, and finally, a linear layer is applied to output a pair of decoded representations. The decoder, shown in Fig. \ref{2}(b), uses the decoupled representations output by the encoder as input to decode the target data. The block structure is illustrated in Fig. \ref{2}(c). The stackable block structure follows the general GPT design and consists of a self-attention module and a feedforward network (FFN) module.
	\section{Experiments}
	\subsection{Dataset}
	We utilized the large resistivity model database (RMD) developed by Asif \textit{et al.} \cite{asif2023dl}, which contains a variety of geologically plausible and geophysically resolvable subsurface structures. The database comprises approximately one million resistivity models, with resistivity values ranging from 1 to 2000 ohm-meters, consisting of 30 layers and a maximum depth of 500 meters. Each model adheres to physical constraints. RMD has been shown to improve performance and generalization while also enhancing the consistency and reliability of deep learning models \cite{liu2024multi}.
	
	We conducted one-dimensional forward modeling \cite{yin2013full} on the model database to obtain the corresponding forward responses. The altitude of the transceiver varies uniformly within the range of 25–100 m above the surface. The ATEM system specifications, including transmit current and receiver configuration, align with those of the AeroTEM IV system described in Bedrosian \textit{et al.} \cite{bedrosian2014airborne}. Additionally, we introduced noise into the forward response data. Part of the noise was derived from actual field measurements, which includes motion-induced noise, nearby or moderately distant sferics noise, cultural and natural electromagnetic noise. As a type of incidental impulse noise, the amount of extraction of the nearby or moderately distant sferics noise was relatively small, while another part was Gaussian noise, as complex field noise distributions are often approximated by a Gaussian distribution \cite{chen2020temdnet}. For the remaining noise, we selected a simulation method that has been shown to closely replicate actual noise \cite{auken2008resolution}, which is defined as:
		\begin{equation}
			s_n=s+N(0,1)[STD^2+(\frac{n_b}{s})^2]^\frac{1}{2}s
		\end{equation}
	Where, \(s_n\) represents the obtained noisy signal, and \(s\) is the forward theoretical data. \(N(0,1)\) denotes the standard Gaussian distribution. STD is the uniform noise, and \(n_b\) is the background noise contribution, which is defined as:
		\begin{equation}
			n_b=b{(\frac{t}{{10}^{-3}})}^{-\frac{1}{2}}
		\end{equation}
	Where, \( b \) is the noise level at 1 ms. It is typically taken between 1 nV/m² and 5 nV/m² \cite{auken2008resolution}. For some models, the forward data and the data after noise addition are shown in Fig. \ref{3}.
	\begin{figure}[!t]
		\centering
		\includegraphics[width=3in]{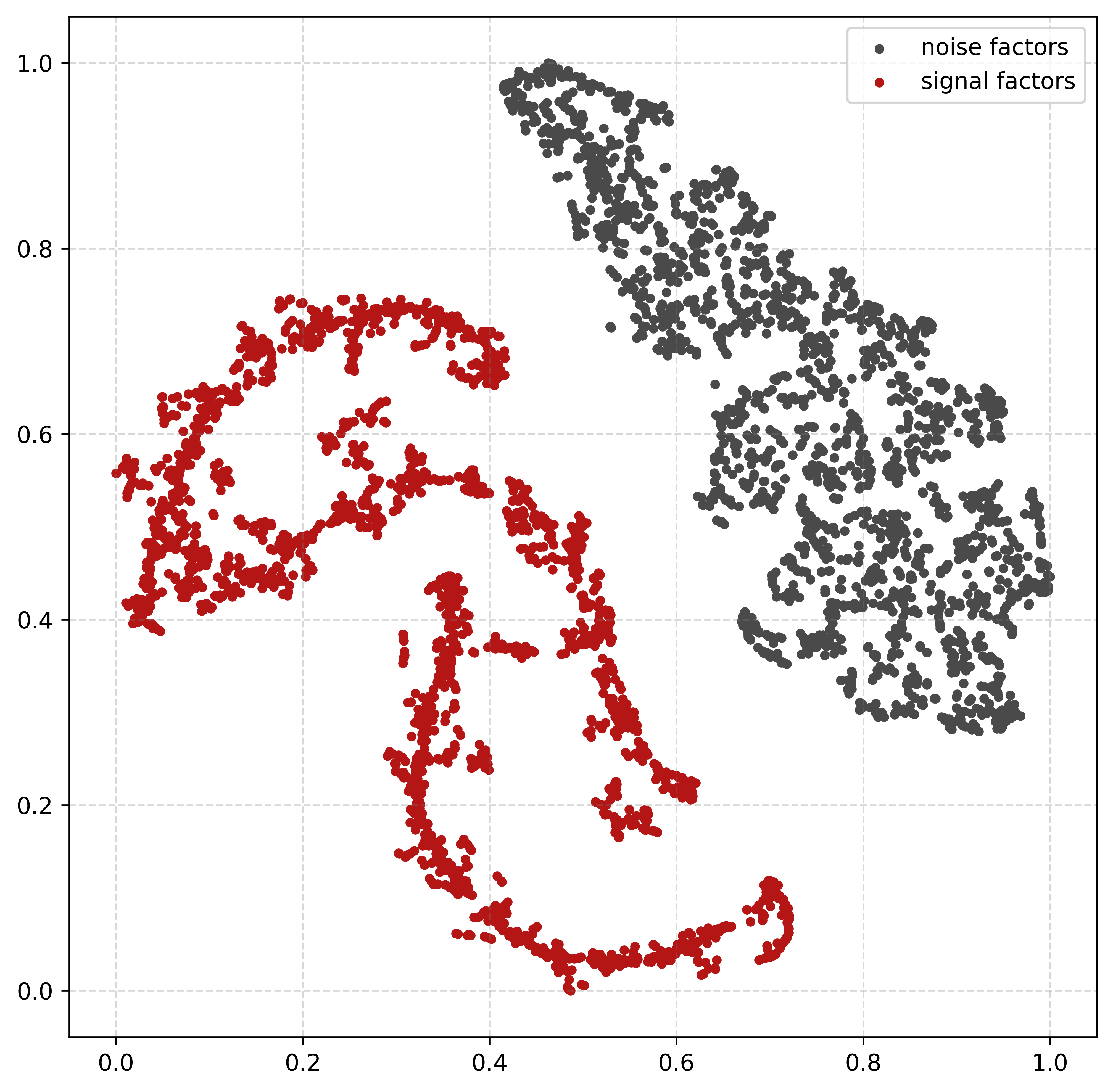}
		\caption{The signal and noise factors are visualized using t-SNE.}
		\label{laten}
	\end{figure}
	\subsection{Train and Evaluation Metric}
	A total of 100,000 samples were generated and randomly divided into training, validation, and test sets in an 8:1:1 ratio, resulting in 10,000 samples each for validation and testing, and 80,000 for training. During training, the number of stacked blocks in both the encoder and decoder was set to 12. The input to the network is time series of shape $B \times 17 \times 1$, where $B$ is the batch size. This input is processed by the embedding layer into latent data of shape $B \times 17 \times 128$. The FFN of blocks consists of two linear layers with a hidden size of 512. The output of the inversion decoder $G_r$ is resistivity model of shape $B \times 30 \times 1$, while the output of the data decoder $G_s$ is time series of shape $B \times 17 \times 1$. Training was performed with a batch size of 32 using the AdamW optimizer, with a learning rate set to 0.0001, for 200 epochs on a single NVIDIA A6000 GPU.
	
	We use the Root Mean Square Percentage Error (RMSPE) to evaluate the accuracy of the inversion model and the signals generated by forward modeling based on this model.
		\begin{equation}
			RMSPE=\sqrt{\frac{1}{M} \sum_{j=1}^M (\frac{p_j^r - p_j^t}{p_j^t})^2 \times 100\%}
		\end{equation}
	Where $p_j^r$ and $p_j^t$ are the predicted values and the true values, respectively; M denotes the number of data points.
	\subsection{Inversion Performance on Test Dataset}
	We first evaluated the performance of our method on the test set, comparing it with the traditional inversion method \cite{yu2018combining} that uses a combined regularization strategy, with conductivity-depth imaging (CDI) \cite{huang2008conductivity} results serving as the initial model. As shown in Fig. \ref{4} (a), when using clean signals for inversion, our method accurately fits the actual model, whereas the traditional approach provides a rough inversion and fails to capture the resistivity variation between 150m and 200m. We then performed forward modeling on the model, with the results shown in Fig. \ref{4} (c). These results demonstrate that the forward signals from our method align better with the actual model’s forward signals, exhibiting smaller errors. Next, we tested the inversion capability on noisy signals. As shown in Fig. \ref{4} (b), since our method decouples the noisy signal into signal and noise factors and uses the signal factors for inversion, it can still accurately decouple the same signal factors and obtain the same inversion result, even in the presence of noise. In contrast, the traditional method lacks adaptability to noisy signals, resulting in larger inversion errors. We performed forward modeling on the inversion results of the noisy signals, with the results shown in Fig. \ref{4} (d). It is evident that our method still fits the actual model’s forward signals well, while the forward signals from the traditional method’s inverted model show significant deviation from the actual model’s forward signals.
	
	To evaluate the stability of the inversion, we analyzed the inversion results for the entire test set, as shown in Fig. \ref{5}. Fig. \ref{5} (a) presents the RMSPE histogram of the inversion results for clean signals using both methods, compared to the true model. Our method demonstrates stable inversion results and accurately predicts the true model, while the traditional method exhibits a wider RMSPE distribution with some extreme values. This indicates that some traditional inversions failed to converge, leading to an increase in RMSPE values. Fig. \ref{5} (b) shows the RMSPE histogram of the forward modeling response of the inversion result in (a) and clean signals. Although the forward modeling response closely matches the original signal values, the traditional method's results still differ from the true model due to the non-uniqueness of ATEM inversion (multiple local minima). The inversion statistical histogram for noisy signals shows that our method maintains accurate inversion results even in the presence of noise, as it decouples the signal factors and performs stable inversion using this factors. In contrast, the traditional method is less robust to noise, resulting in significant deviation from the true model.
	\begin{figure*}[!t]
		\centering
		\includegraphics[width=7in]{./Fig/height}
		\caption{Inversion results at different heights: (a–d) correspond to heights of 30, 50, 70, and 90 m, respectively, while (e–h) show the forward responses obtained from the corresponding inversion results.}
		\label{height}
	\end{figure*}
	
	To ensure the network correctly decouples the data and that the encoder $E$ accurately encodes the data into signal and noise factors, we analyzed a set of clean and noisy signals and conducted statistical analysis on the test set, as shown in Fig. \ref{6}. The clean signal \( s \) was decoupled into the signal factors \( Z_s^1 \) and the noise factors \( Z_n^1 \), while the noisy signal \( n \) was decoupled into the signal factors \( Z_s^2 \) and the noise factors \( Z_n^2 \). By using various combinations of signal and noise factors, we were able to accurately reconstruct the corresponding signals, confirming that the network effectively decouples the data. We also performed a statistical analysis of the RMSPE histograms of the data decoded using \( (Z_s^1, Z_n^1) \) and \( (Z_s^2, Z_n^1) \), compared to the clean signal, as shown in Fig. \ref{6} (e) and (f). These results highlight the network’s stable decoupling capability, demonstrating its ability to correctly encode the data into signal and noise factors.
	
	We visualized the signal and noise factors to demonstrate the interpretability of the proposed method. The visualization was performed using t-SNE \cite{maaten2008visualizing}, which projects high-dimensional latent representations into a low-dimensional space, allowing for an intuitive observation of data distances and distributions. As shown in Fig. \ref{laten}, the signal and noise factors exhibit distinct, non-overlapping distributions, indicating that the network effectively disentangles the data. Moreover, since the inversion results are derived from the signal factors, the interpretability of the network is further enhanced.

	\subsection{Sensitivity to transceiver height}
	To evaluate the network’s sensitivity to variations in height, we randomly selected a model from the test set and performed forward modeling with heights set to 30, 50, 70, and 90 meters, respectively. The resulting data were then inverted using the proposed network, as illustrated in the Fig. \ref{height}. It can be observed that the amplitude of the AEM signals varies with height, yet the network effectively handles such height-induced changes. Under different height conditions, the network consistently produces satisfactory inversion results.
	
	\subsection{Field Data Application}
\begin{figure*}[!t]
		\centering
		\includegraphics[width=6in]{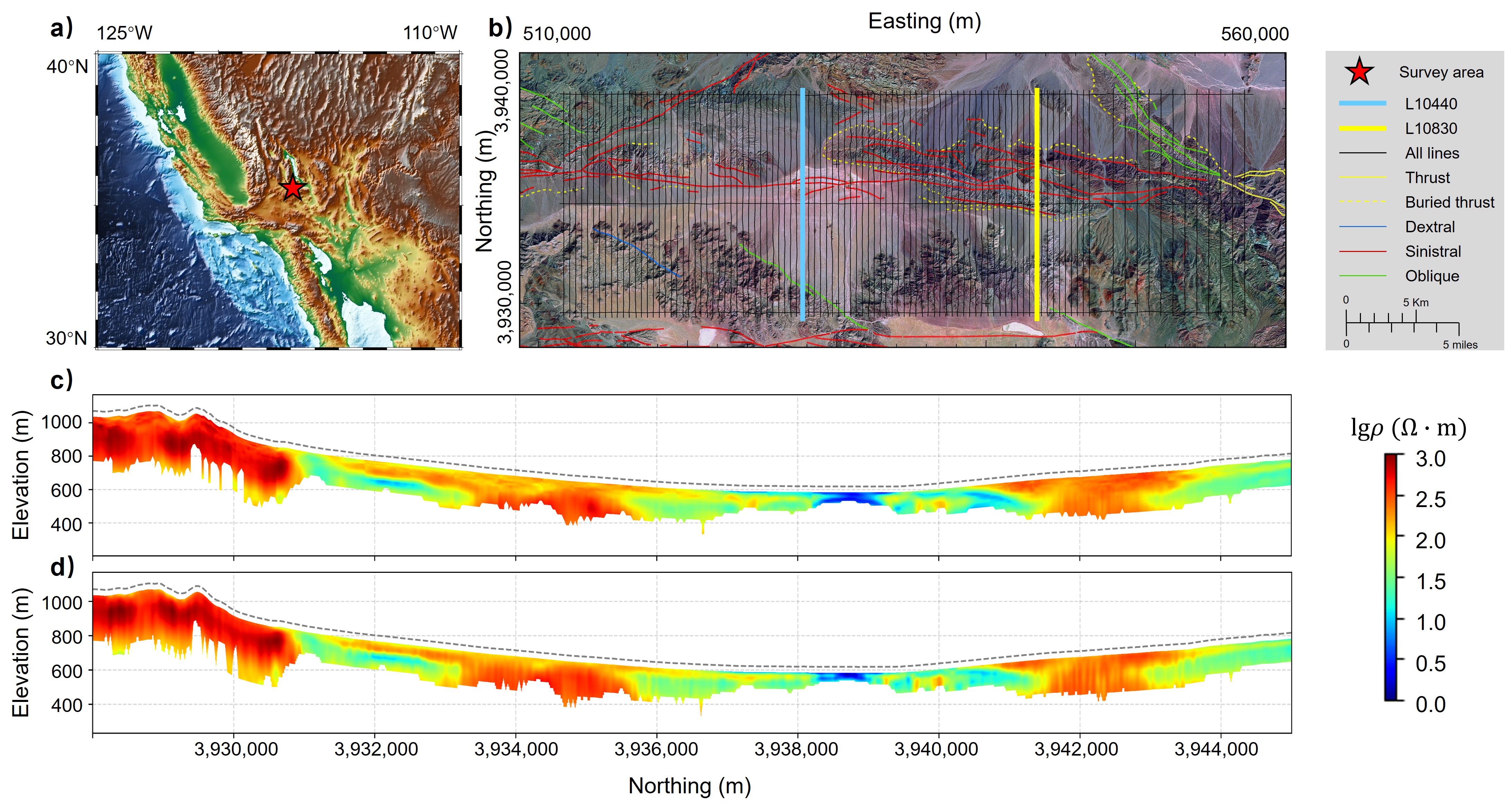}
		\caption{(a) Survey area, marked with a pentagram. (b) Distribution of flight lines (thin black lines). The thick blue line represents the L10440 survey line. (c) Resistivity profile of the L10440 survey line obtained by the USGS using traditional methods, consisting of 535 resistivity models. The black dashed line indicates the transceiver height. (d) Resistivity profile of the L10440 survey line obtained using our method, consisting of 6423 resistivity models.}
		\label{9}
	\end{figure*}

	To demonstrate the applicability of the method, we tested the trained model on ATEM data collected by the United States Geological Survey (USGS) in the Leach Lake Basin at Fort Irwin, California, USA (Fig. \ref{9} (a)). This survey area is a geologically complex, internally drained basin, characterized by numerous faults on both sides. The total length of the flight lines is 1,785.95 km, encompassing over 740,000 ATEM time series, as shown in Fig. \ref{9} (b). Further details on the data can be found in the report by the USGS \cite{bedrosian2014airborne}.
	
	We present the inversion results for the L10440 survey line in Fig. \ref{9}. The USGS excluded data that could not be inverted due to severe noise interference and downsampled the remaining data by a factor of 10, resulting in 535 19-layer models, as shown in Fig. \ref{9} (c). In contrast, our method successfully processed all 6,423 data points, yielding 6,423 30-layer models, as shown in Fig. \ref{9} (d). Our method effectively reveals the subsurface electrical structure, aligning with the results obtained by the USGS. It clearly identifies faults, low-resistance areas, and high-resistance regions, demonstrating the accuracy and reliability of our inversion approach.
	
\begin{figure*}[!t]
		\centering
		\includegraphics[width=6in]{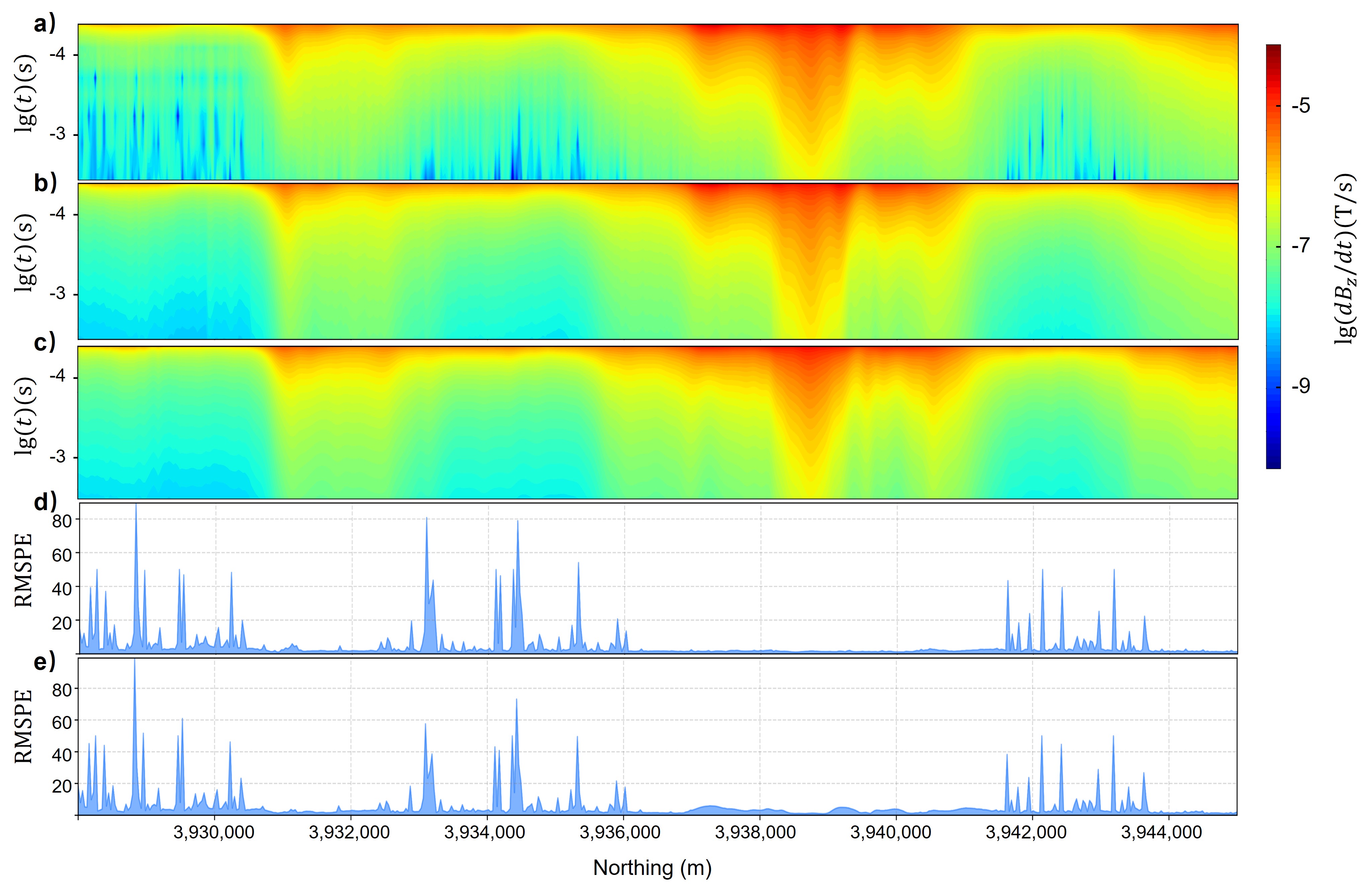}
		\caption{(a) Original data along Line L10440. (b) Forward response obtained using the USGS inversion results. (c) Forward response obtained using the inversion results from our method. (d) RMSPE between (b) and (a). (e) RMSPE between (c) and (a).}
		\label{10}
	\end{figure*}

	To further evaluate the inversion results, forward modeling was performed using the inverted one-dimensional models, and the simulated data were compared with the original observations, as shown in Fig. \ref{10}. It can be seen that in regions with low noise levels, the RMSPE is small, and the forward-modeled data closely match the observed data. In contrast, in regions with high noise levels, the presence of significant uncertainty introduced by the noise leads to large spikes in RMSPE and substantially higher values.
	
\begin{figure*}[!t]
		\centering
		\includegraphics[width=6in]{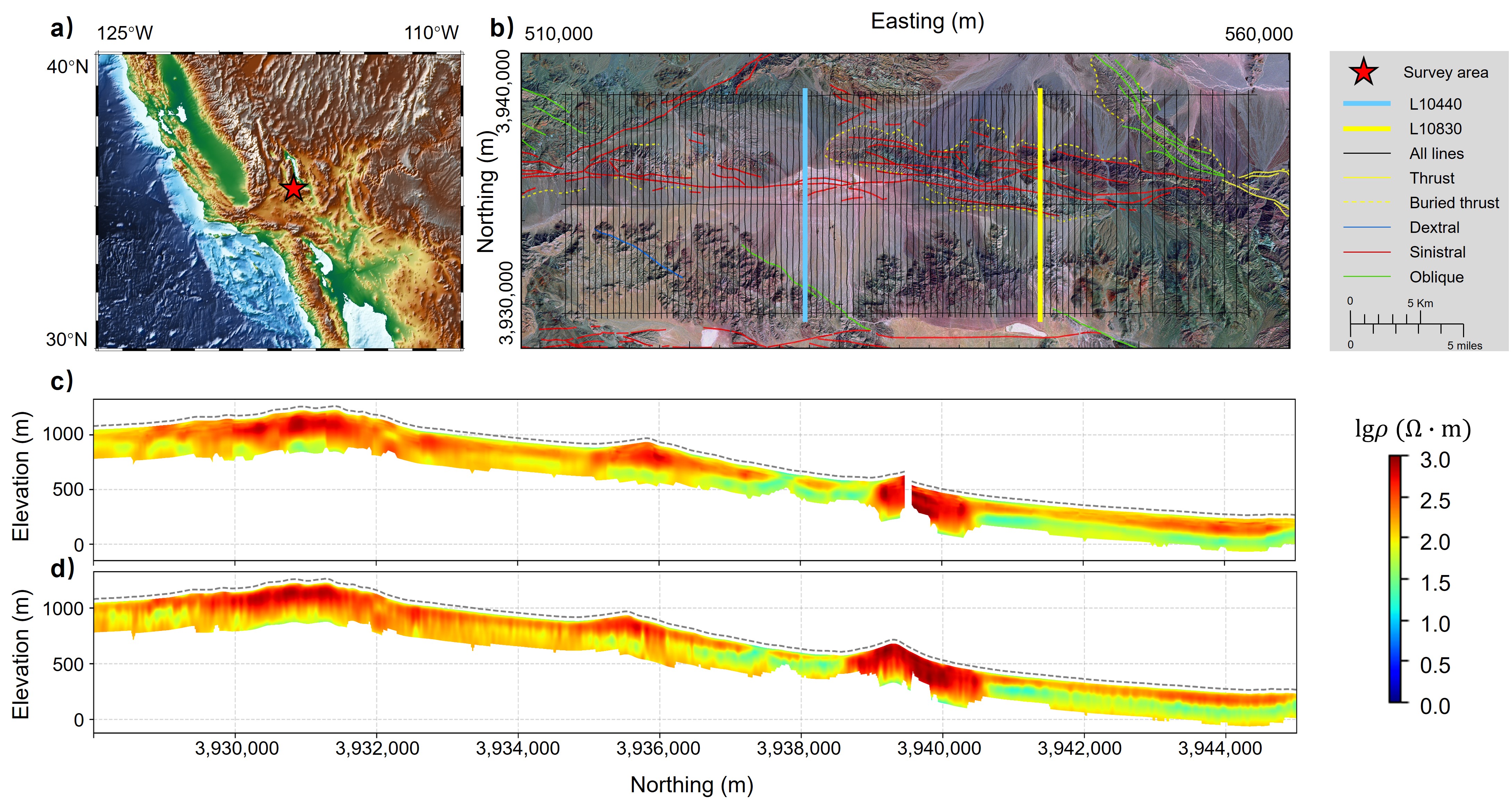}
		\caption{(a) Survey area, marked with a star. (b) Distribution of flight lines (thin black lines). The thick yellow line represents the L10830 survey line. (c) The resistivity profile of the L10830 survey line obtained by the USGS using traditional methods, consisting of 515 resistivity models. The black dashed line indicates the transmitter-receiver height. (d) The resistivity profile of the L10830 survey line obtained by our method, consisting of 6412 resistivity models.}
		\label{11}
	\end{figure*}

	We present the inversion results for the L10830 survey line in Fig. \ref{11}. The USGS excluded data significantly affected by environmental noise from their 515 19-layer inversion results, which led to gaps in the inversion profile and impacted the lateral resolution of the subsurface electrical structure. In contrast, our method processes all 6,412 data points without being influenced by noise, yielding 6,412 30-layer models, as shown in Fig. \ref{11} (d). Our method provides subsurface electrical structures consistent with those obtained by the USGS, clearly identifying faults and other stratigraphic layers, while also revealing the gaps in the USGS inversion results, offering improved lateral resolution.
	\begin{figure*}[!t]
		\centering
		\includegraphics[width=6in]{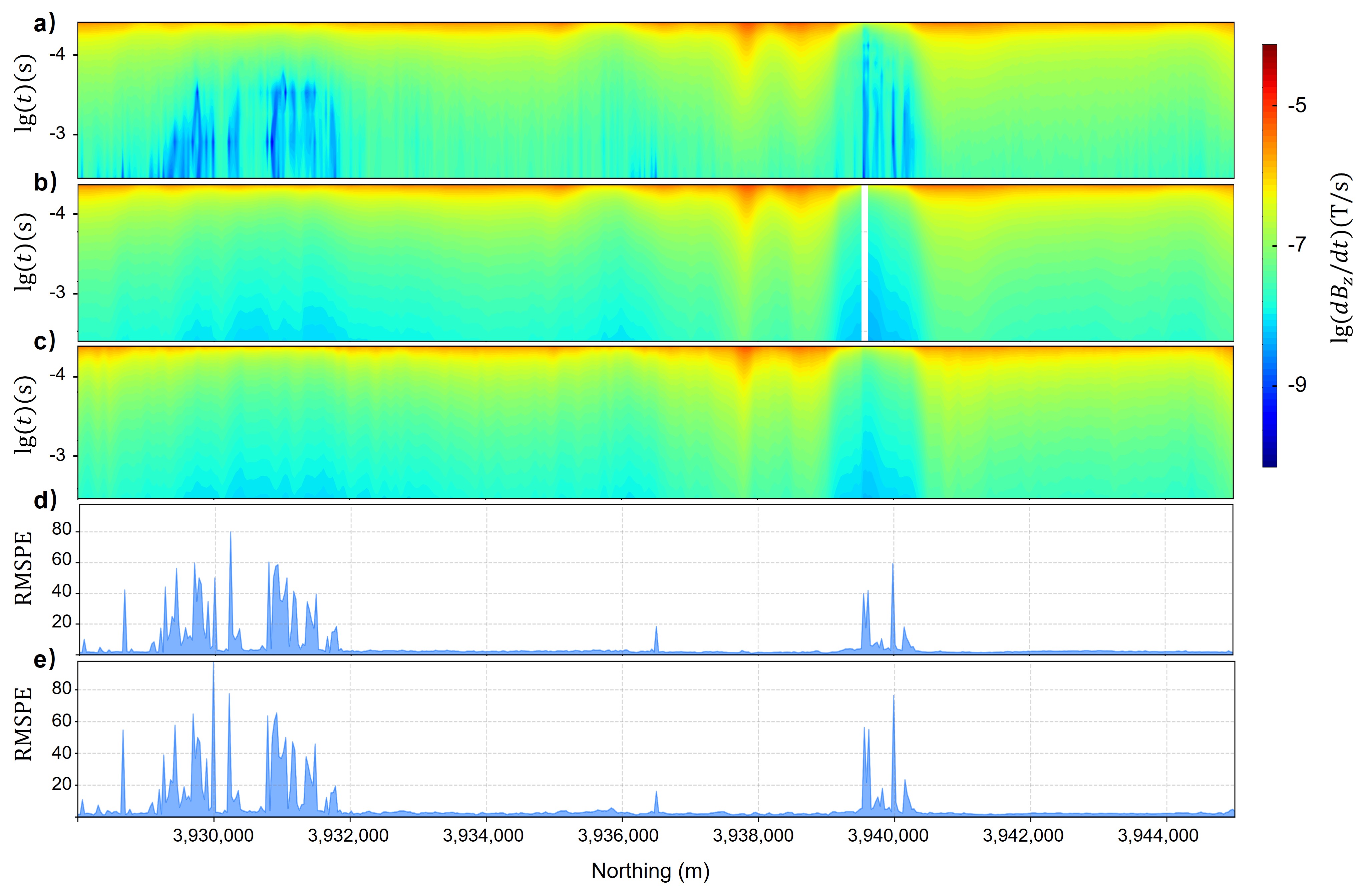}
		\caption{(a) Original data along Line L10830. (b) Forward response obtained using the USGS inversion results. (c) Forward response obtained using the inversion results from our method. (d) RMSPE between (b) and (a). (e) RMSPE between (c) and (a).}
		\label{12}
	\end{figure*}
	
	Forward modeling was performed using the inverted one-dimensional models, and the results were compared with the original observations, as shown in Fig. \ref{12}. The AEM signals predicted by our method are generally consistent with both the observed data and the signals predicted by the USGS reconstructed model. It is important to note that, in real-world scenarios, noise-free AEM data are not available. Thus, the RMSPE is approximately estimated based on the original observed data, which contain noise, and therefore may not approach zero.
	
\begin{figure*}[!t]
		\centering
		\includegraphics[width=6in]{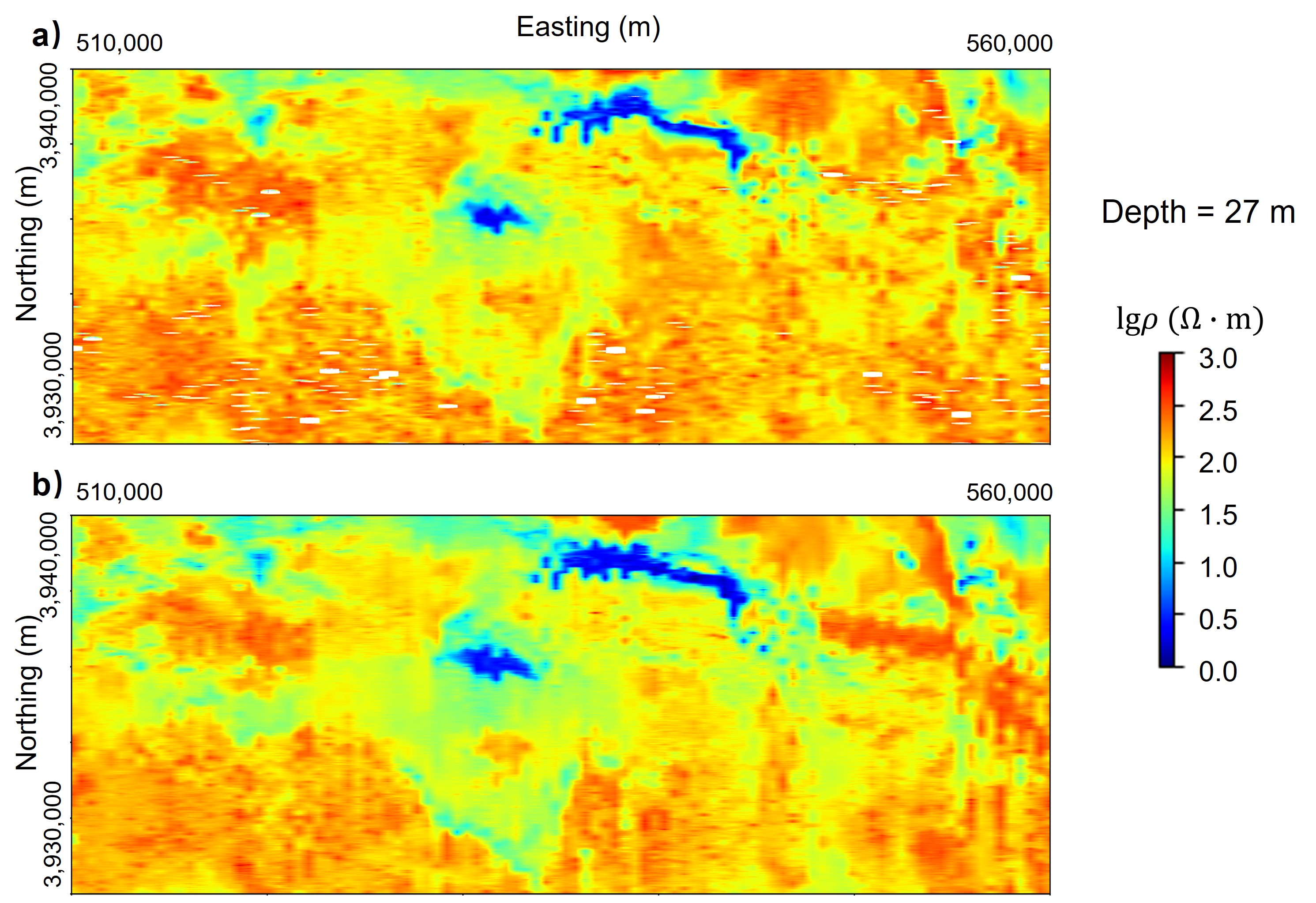}
		\caption{Resistivity slices at a depth of 27 meters. (a) Inversion results obtained by USGS using traditional methods. (b) Inversion results obtained by our method.}
		\label{13}
	\end{figure*}

	To validate the depth results obtained by the proposed method, we generated depth slices to analyze the lateral variations in resistivity. The depth slice at a depth of 27 meters is shown in Fig. \ref{13}. The results from our method are highly consistent with the USGS inversion results, clearly delineating the edge and extent of Leach Lake and the surrounding mountain contours. In contrast, the USGS inversion results excluded resistivity models with low reliability that were significantly affected by environmental noise, leading to gaps in the depth slice and affecting the lateral resolution of the subsurface electrical structure. Our method, however, processes all the data, providing a more accurate lateral representation.
	
\begin{figure*}[!t]
		\centering
		\includegraphics[width=6in]{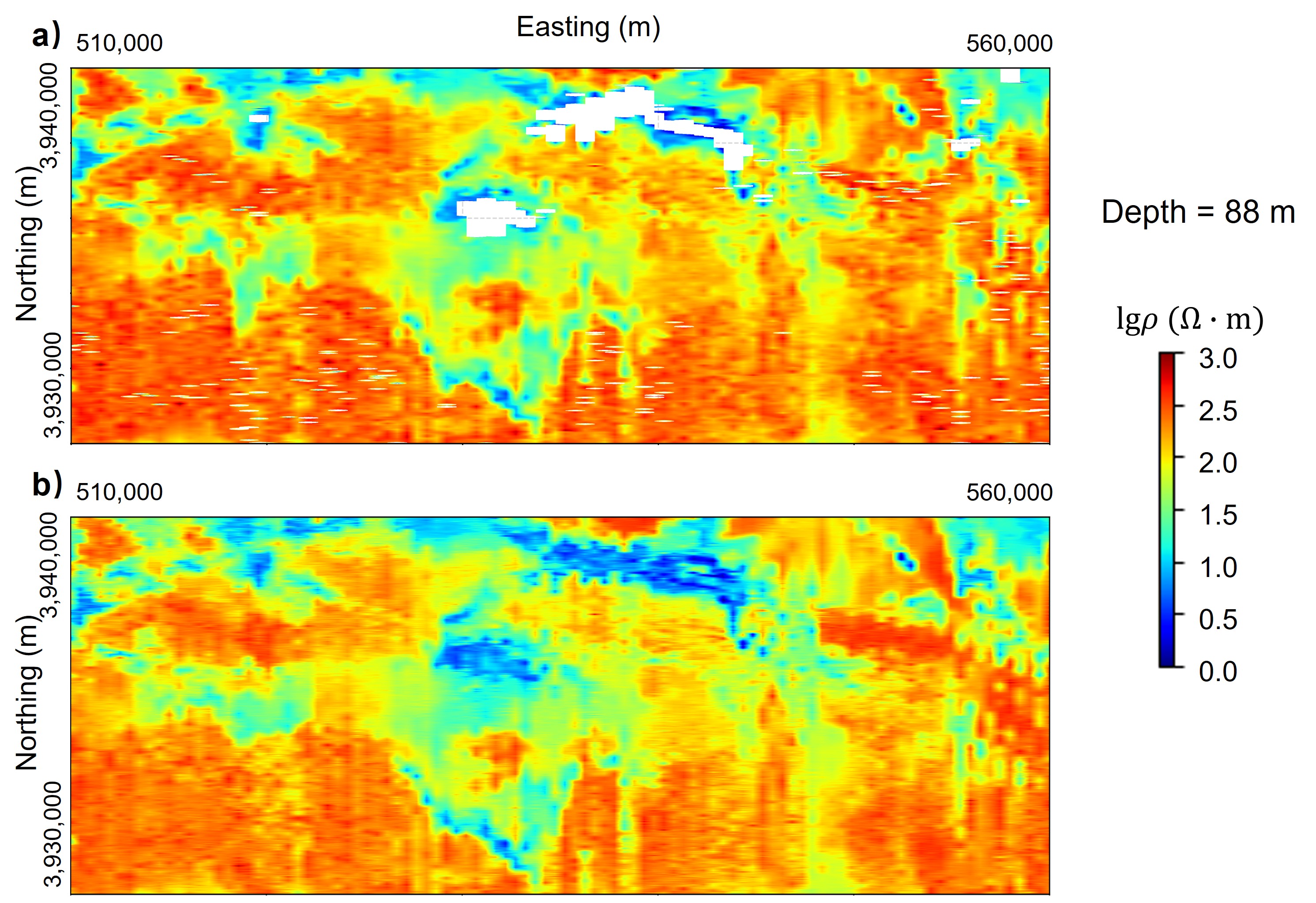}
		\caption{Resistivity slices at a depth of 88 meters. (a) Inversion results obtained by USGS using traditional methods. (b) Inversion results obtained by our method.}
		\label{14}
	\end{figure*}

	The depth slice at 88 meters, shown in Fig. \ref{14}, demonstrates that the results obtained by our method are consistent with the USGS inversion results, revealing well-defined lateral structures. This indicates that the inversion results produced by our proposed method are reliable at this depth and can be trusted for accurate interpretation.
	
	The inversion results of field data from the Leach Lake Basin at Fort Irwin demonstrate that our method yields reliable results, accurately revealing the subsurface electrical structure. Additionally, it can directly handle noisy data, including data affected by significant environmental noise, which traditional methods struggle to process, providing enhanced lateral structural representations. Furthermore, the RMD used, with its well-structured and deep layers, enables the trained deep learning model to accurately reflect the actual subsurface electrical structure.
	\section{Discussion}
	In this study, we proposed an interpretable deep learning inversion method for AEM, which leverages disentangled representation learning to separate noisy data into noise and signal factors. The inversion is then performed using the signal factors, thereby enhancing both the interpretability and reliability of the results. 
	
	The dataset generation took 1.6 hours, and each training epoch required around 3.18 minutes, resulting in a total training time of about 10.6 hours. After training, inference on all 6,423 data points along L10440 took 0.53 seconds and inference on all 6,412 data points along L10830 took 0.52 seconds. In total, the entire process took approximately 12.2 hours. In contrast, traditional methods required about 14.6 hours to process Lines 10440 and 10830. Even when accounting for data preparation and training time, our method is more efficient than traditional approaches. Therefore, when applied to the full set of observational data across the survey area, our method demonstrates superior efficiency.
	
	When switching to a different survey system, a small-scale transfer learning process is required to allow the model to adapt to the new system. However, this adaptation can be performed in advance and only needs to be done once for each given system. Overall, the inversion efficiency remains significantly higher than that of traditional inversion methods.
	
	\section{Conclusions}
	In this study, we propose a unified, interpretable deep learning inversion framework based on decoupled representation learning. The network effectively decouples noisy data into signal and noise factors, utilizing the signal factors for inversion. This approach integrates denoising and inversion processes for field data. By incorporating physical information as guidance and leveraging signal factors throughout the data processing pipeline, the network are more reliable and interpretable. The inversion results from models trained on the RMD using field data demonstrate that our model can directly derive accurate subsurface electrical structures from noisy data. Moreover, it is capable of processing data affected by significant environmental noise, which traditional methods cannot handle, offering enhanced lateral structural representations.	

\normalem
\bibliography{r}

\bibliographystyle{IEEEtran}
		
\end{document}